\documentclass[]{spie}  %>>> use for US letter paper

\usepackage{amsmath,amsfonts,amssymb}
\usepackage{subcaption}
\usepackage{graphicx}
\usepackage{cite} 
\usepackage{epsfig}
\usepackage{nccmath}
\usepackage{xcolor,colortbl}
\usepackage{tabularx}
\usepackage{relsize}
\usepackage{pifont}
\usepackage{booktabs} 
\usepackage{multirow}
\usepackage{multicol}
\usepackage{adjustbox}
\usepackage{float}
\usepackage{makecell}
\usepackage{tabu}
\usepackage[colorlinks=true, allcolors=blue]{hyperref}
\usepackage[capitalize]{cleveref}
\usepackage{algorithm}
\usepackage{adjustbox}

\title{Comprehensive Evaluation and Fine-Tuning of Foundational Cell Nuclei Segmentation Models in Renal Pathology }

\author[a$\dag$]{Ruijie Wu}
\author[a$\dag$]{Junlin Guo}
\author[b]{Ruining Deng}
\author[c]{Yu Wang}
\author[c]{Shilin Zhao}
\author[d]{Haichun Yang}
\author[a,d,e]{Yuankai Huo}

\affil[a]{Department of Electrical and Computer Engineering, Vanderbilt University, Nashville, TN 37235, USA}
\affil[b]{Department of Radiology, Weill Cornell Medicine, New York, NY 10021, USA}
\affil[c]{Department of Biostatistics, Vanderbilt University Medical Center, Nashville, TN 37235, USA}
\affil[d]{Department of Pathology, Microbiology and Immunology, Vanderbilt University Medical Center, Nashville, TN 37235, USA}
\affil[e]{Department of Computer Science, Vanderbilt University, Nashville, TN 37235, USA}

\affil[$\dag$]{These authors contribute equally to this work.}

\authorinfo{Further author information: (Send correspondence to Yuankai Huo)\\
E-mail: yuankai.huo@vanderbilt.edu}

\begin{document}
\maketitle

\begin{abstract}

Accurate nuclei instance segmentation is essential for quantitative renal pathology, yet general-purpose AI foundation models often struggle with renal tissue due to low contrast, densely packed nuclei, complex morphology, and strong background staining. Consequently, domain-specific fine-tuning is often necessary to bridge the gap between general-purpose segmentation capability and reliable performance in focused pathology applications. In our prior SPIE Medical Imaging studies, we demonstrated a substantial performance gap when applying multiple general-purpose cell segmentation foundation models to renal pathology. In our other study, we further showed that fine-tuning with well-segmented ``Easy" cases and model-failure ``Hard" cases could substantially improve segmentation performance. However, an important piece remained unresolved: whether annotating the much larger and more time-consuming set of intermediate-difficulty cases provides additional value beyond Easy and Hard examples. In this study, we complete this evaluation by introducing a ``Medium"-difficulty cohort and systematically assessing its contribution to model adaptation. Specifically, we extended our human-in-the-loop framework by combining 5,901 foundation-model-generated pseudo-labels from well-segmented cases (Easy), 860 newly expert-annotated unresolved challenging cases (Medium), and 198 expert-annotated cases that were jointly identified as failures by the foundation models in our prior study (Hard). We additionally incorporated recently developed general-purpose cell segmentation models to evaluate the performance of current state-of-the-art approaches. Together, these annotations spanning multiple levels of segmentation difficulty enabled a systematic evaluation of seven single-source and mixed-source fine-tuning strategies across nine cell segmentation model configurations. The segmentation results demonstrate that annotations across different levels of segmentation difficulty can effectively support domain adaptation, while also revealing that the optimal composition of fine-tuning data remains model dependent. The code and resources used in this study are publicly available at 
\url{https://github.com/hrlblab/AFM_kidney_cells/tree/main/stage4_paper}.

\end{abstract}

\section{Introduction}
Renal pathology presents a particularly demanding setting for cell nuclei instance segmentation. Although general-purpose cell segmentation models have improved considerably, their performance remains inconsistent in kidney-specific regions that differ from the images used for pretraining, and architectural advances alone do not guarantee that difficult kidney nuclei will be adequately segmented. Samples with low contrast, unusual morphology, dense nuclear distributions, or strong background staining may remain unresolved even after applying newer models~\cite{guo2025assessment,guo2025evaluating,wang2026evaluating}. These limitations motivate domain-specific adaptation of general-purpose segmentation models using kidney-specific supervision.

Our previous studies~\cite{guo2025assessment,guo2025evaluating,wang2026evaluating} progressively identified this domain gap and explored two complementary directions for addressing it: human-in-the-loop data enrichment and advances in general-purpose cell segmentation models. In our prior study (2024)~\cite{guo2025assessment}, 
Cellpose~\cite{stringer2021cellpose}, 
StarDist~\cite{weigert2022stardist}, 
and CellViT~\cite{horst2024cellvit} 
were evaluated on approximately 8,800 kidney pathology image patches; 
even the best-performing model, CellViT, adequately segmented 63.6\% of images, 
leaving approximately 37\% insufficiently resolved. In 2025, a human-in-the-loop data-enrichment strategy using scalable Easy pseudo-labels and 198 expert-annotated Hard cases improved all three models on kidney pathology instance segmentation~\cite{guo2025evaluating}. More recently, newly released models converted 1,301 of 2,091 previously challenging patches to Good-quality predictions, while 782 remained Medium and eight remained Bad; across the full evaluation set, 1,079 Medium-quality patches remained and formed the candidate pool for annotation.\cite{wang2026evaluating} These remaining Medium cases represented a substantially larger unresolved cohort than the 198 previously annotated Hard cases. Unlike scalable Easy pseudo-labels and the relatively small set of consensus Hard failures, incorporating these Medium cases would require substantial additional expert annotation, while the benefit of this annotation effort remained unclear. Therefore, a key question arises: \textbf{Does expert annotation of the remaining Medium-rated cases provide additional value beyond existing Easy pseudo-labels and Hard expert annotations, and is this benefit consistent across different segmentation models?}

\begin{figure}[ht]
    \centering
    \includegraphics[width=1\textwidth]{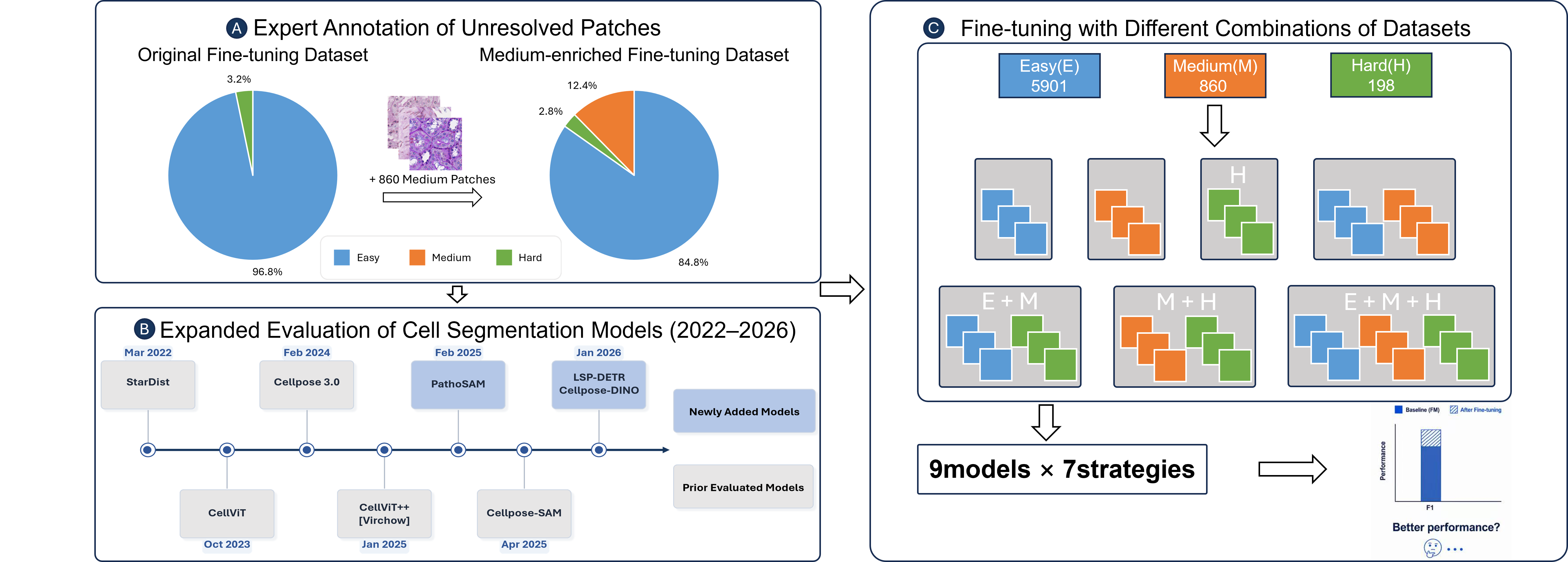}
    \caption{\textbf{Overall framework.} (A) Expert annotation of previously unresolved Medium-quality kidney pathology patches to enrich the existing Easy and Hard fine-tuning data. (B) Expanded evaluation of cell nuclei segmentation models released from 2022 to 2026, including both previously evaluated and newly incorporated models. (C) Fine-tuning with seven single-source and mixed-source combinations of Easy, Medium, and Hard datasets to evaluate model-specific adaptation performance.}
    \label{fig:fig1}
\end{figure}

Motivated by this remaining gap, we extend our previous human-in-the-loop framework by introducing expert-annotated Medium cases, enabling a systematic evaluation across Easy, Medium, and Hard annotation sources.
We use 5,901 foundation-model-generated pseudo-labels from well-segmented cases as Easy data and 198 prior expert-annotated consensus failure cases as Hard data~\cite{guo2025evaluating}.
Pathology experts selected and annotated 860 challenging patches from a pool of 1,079 previously unresolved Medium-quality image patches (Fig.~\ref{fig:fig1}A).
Together, these annotation sources were used to evaluate whether incorporating Medium annotations provides additional value beyond the previously established Easy + Hard fine-tuning strategy for kidney nuclei instance segmentation.~\cite{guo2025evaluating,wang2026evaluating}.
We fine-tuned and evaluated nine cell segmentation model configurations released from 2022 to 2026, including Cellpose 3.0, Cellpose-SAM, Cellpose-DINO, StarDist, CellViT (HIPT and SAM), CellViT++ (Virchow), LSP-DETR, and PathoSAM~\cite{stringer2025cellpose3,pachitariu2025cellposesam,cellpose2026docs,simeoni2025dinov3,weigert2022stardist,horst2024cellvit,horst2026cellvitpp,pekar2026lspdetr,griebel2025pathosam} (Fig.~\ref{fig:fig1}B).
Each model was fine-tuned using seven single-source and mixed-source strategies across Easy, Medium, and Hard annotation sources, resulting in 63 fine-tuning experiments (Fig.~\ref{fig:fig1}C).
The overall framework for this study is shown in Fig.~\ref{fig:fig1}.

\noindent {The main contributions of our study are as follows:}  
\begin{itemize}  
    \item \textbf{Expert annotation of previously unresolved Medium patches}: We extend human-in-the-loop enrichment beyond Easy pseudo-labels and Hard failure cases by selecting and annotating Medium-quality patches that remained insufficiently segmented after successive model evaluations. 

    \item \textbf{Import recent cell nuclei segmentation models released in 2025–2026}: We systematically evaluate nine cell segmentation model configurations under the same annotation enrichment framework and show that the optimal composition of Easy, Medium, and Hard annotations varies across models.

    \item \textbf{Fine-tuning with different combinations of Easy, Medium and Hard datasets}: We systematically evaluate Easy pseudo-labels, Medium annotations, and Hard expert annotations, both separately and in different combinations, to determine whether Medium annotations provide additional benefits when combined with large-scale pseudo-labels, challenging expert-labeled data, or both.

\end{itemize}

% In this paper, we propose a universal medical dataset distillation pipeline for effective data sharing in healthcare, and we provide comprehensive experimental analysis methods to address the above three key questions. This paper will contribute to the following parts:

% $\bullet$ Assess the impact of data distillation across multiple datasets characterized by minor or great variations.

% $\bullet$ Explore the indicator to predict the distillation performance. 

\section{Methods}

% In this study, the overall framework includes 3 steps. (1) Select the most challenging kidney nuclei image patches and refined instance annotations through a human-in-the-loop process. (2) Construct diverse datasets with Easy pseudo-labels and prior Hard cases. (3) Design different fine-tuning strategies for fine-tuning.

\subsection{Comprehensive Cell Nuclei Segmentation Model Benchmark}
In this study, we chose nine cell segmentation model configurations to fine-tune, spanning models released from 2022 to 2026, as illustrated in Fig.~\ref{fig:fig1}B and summarized in Table~\ref{tab:cell-foundation-models}. The models can be categorized into three groups. The first group consists of Cellpose 3.0, StarDist, and CellViT, which were fine-tuned in our previous study and were re-fine-tuned here using the new dataset~\cite{guo2025evaluating}. The second group, CellViT++ [Virchow] and Cellpose-SAM, had previously been evaluated only for inference~\cite{wang2026evaluating}. The third group, PathoSAM, LSP-DETR, and Cellpose-DINO, comprises models that had not been evaluated or fine-tuned in our prior kidney studies~\cite{griebel2025pathosam,pekar2026lspdetr,cellpose2026docs}.

\begin{enumerate}
    % \item \textbf{Cellpose-based models}: Cellpose 3.0, Cellpose-SAM, and Cellpose-DINO follow the Cellpose flow-based instance segmentation framework. They predict horizontal and vertical pixel flows together with a cell-probability map and reconstruct individual objects by grouping pixels that converge to the same location. Cellpose-SAM and Cellpose-DINO replace the convolutional encoder with pretrained SAM and DINOv3 ViT encoders, respectively, while retaining the Cellpose prediction and instance-reconstruction framework~\cite{stringer2021cellpose,stringer2025cellpose3,pachitariu2025cellposesam,cellpose2026docs,simeoni2025dinov3}.

    \item \textbf{Cellpose-based models}: Cellpose 3.0, Cellpose-SAM, and Cellpose-DINO follow the Cellpose flow-based instance segmentation framework. They predict horizontal and vertical pixel flows together with a cell-probability map and reconstruct individual objects by grouping pixels that converge to the same location. Cellpose-SAM and Cellpose-DINO replace the convolutional encoder with pretrained SAM and DINOv3 ViT encoders, respectively, while retaining the Cellpose prediction and instance-reconstruction framework~\cite{stringer2021cellpose,stringer2025cellpose3,pachitariu2025cellposesam,cellpose2026docs,simeoni2025dinov3}. For Cellpose 3.0, we followed our prior preprocessing pipeline by converting RGB histopathology patches into grayscale, DAPI-like nuclear inputs for both fine-tuning and inference~\cite{guo2025evaluating}.

    \item \textbf{StarDist}: StarDist represents each nucleus as a star-convex polygon defined by radial distances from an interior pixel to the object boundary. The network densely predicts object probabilities and radial distances along predefined directions. The resulting polygon candidates are filtered using non-maximum suppression to obtain the final nucleus instances~\cite{weigert2022stardist}.

    \item \textbf{CellViT-based models}: CellViT combines a pretrained ViT encoder with a U-Net-like decoder for nucleus instance segmentation. It predicts a binary nuclei map, horizontal and vertical distance maps, and a nuclei-type map, followed by HoVer-Net-style post-processing to separate touching nuclei~\cite{horst2024cellvit,graham2019hovernet}. The HIPT and SAM variants primarily differ in their pretrained encoders~\cite{chen2022hipt,kirillov2023sam}. CellViT++ retains the CellViT segmentation framework while incorporating foundation-model encoders, including the pathology-specific Virchow encoder, and nucleus-level embeddings for downstream cell classification~\cite{horst2026cellvitpp,vorontsov2024virchow}.

    \item \textbf{LSP-DETR}: LSP-DETR formulates nucleus instance segmentation as a direct set-prediction problem. Each nucleus is represented by a center point and radial distances defining a star-convex boundary. A DETR-style decoder predicts nucleus locations and shapes, allowing instance masks to be generated directly without conventional watershed-based post-processing~\cite{carion2020detr,pekar2026lspdetr}.

    \item \textbf{PathoSAM}: PathoSAM adapts SAM to histopathology images for both interactive and automatic instance segmentation. Its automatic instance-segmentation decoder predicts foreground probabilities, distances to the closest object centers, and distances to the closest object boundaries. These prediction maps are then processed using seeded watershed to separate adjacent nuclei and generate individual instances~\cite{kirillov2023sam,griebel2025pathosam}.

\end{enumerate}

% \begin{table}[h!]
% \caption{SOTA Cell Segmentation Foundation models in 2025}
% \label{tab:my-table}
% \begin{tabular}{@{}cccc@{}}
% \toprule
% Model                   & Year-Month & Backbone                      & Segmentation Approach                                     \\ \midrule
% CellViT++ {[}SAM{]}     & 2025 Jan   & SAM-H & ViT encoder + segmentation decoder                        \\ \midrule
% CellViT++ {[}HIPT{]}    & 2025 Jan   & HIPT-256                      & ViT encoder + segmentation decoder                        \\ \midrule
% CellViT++ {[}Virchow{]} & 2025 Jan   & Virchow                       & ViT encoder + segmentation decoder                        \\ \midrule
% Cellpose-SAM            & 2025 Apr   & U-Net + ViT (SAM)             & ViT encoder + flow field prediction and gradient tracking \\ \bottomrule
% \end{tabular}
% \end{table}

\begin{table}[ht]
\centering
\caption{Cell Nuclei Segmentation Models in Pathology (2022-2026)}
\label{tab:cell-foundation-models}
\begin{adjustbox}{max width=\textwidth}
\begin{tabular}{@{}lllc@{}}
\toprule
\textbf{Models}             & \textbf{Release Date} & \textbf{Backbone}          & \textbf{Code}                                  \\ \midrule
StarDist (Histo.)~\cite{weigert2022stardist} & 2022 Mar & U-Net~\cite{ronneberger2015unet} & \href{https://github.com/stardist/stardist}{URL} \\
CellViT [HIPT-256]~\cite{horst2024cellvit} & 2023 Oct & HIPT ViT~\cite{chen2022hipt} & \href{https://github.com/TIO-IKIM/CellViT}{URL} \\
CellViT [SAM-H]~\cite{horst2024cellvit} & 2023 Oct & SAM ViT~\cite{kirillov2023sam} & \href{https://github.com/TIO-IKIM/CellViT}{URL} \\
Cellpose 3.0~\cite{stringer2025cellpose3} & 2024 Feb & U-Net~\cite{ronneberger2015unet} & \href{https://github.com/MouseLand/cellpose}{URL} \\
CellViT++ [Virchow]~\cite{horst2026cellvitpp} & 2025 Jan & Virchow ViT~\cite{vorontsov2024virchow} & \href{https://github.com/TIO-IKIM/CellViT-plus-plus}{URL} \\
PathoSAM~\cite{griebel2025pathosam} & 2025 Feb & SAM ViT~\cite{kirillov2023sam} & \href{https://github.com/computational-cell-analytics/patho-sam}{URL} \\
Cellpose-SAM~\cite{pachitariu2025cellposesam} & 2025 Apr & SAM ViT~\cite{kirillov2023sam} & \href{https://github.com/MouseLand/cellpose}{URL} \\
LSP-DETR~\cite{pekar2026lspdetr} & 2026 Jan & SwinV2-T~\cite{liu2022swinv2} & \href{https://github.com/RationAI/lsp-detr}{URL} \\
Cellpose-DINO~\cite{cellpose2026docs} & 2026 Jun & DINOv3 ViT~\cite{simeoni2025dinov3} & \href{https://github.com/MouseLand/cellpose}{URL} \\
\bottomrule
\end{tabular}
\end{adjustbox}
\end{table}

\subsection{Human-in-the-Loop Annotation and Multi-source Data Enrichment}
\begin{figure}[ht] 
    \centering 
    \includegraphics[width=1\textwidth]{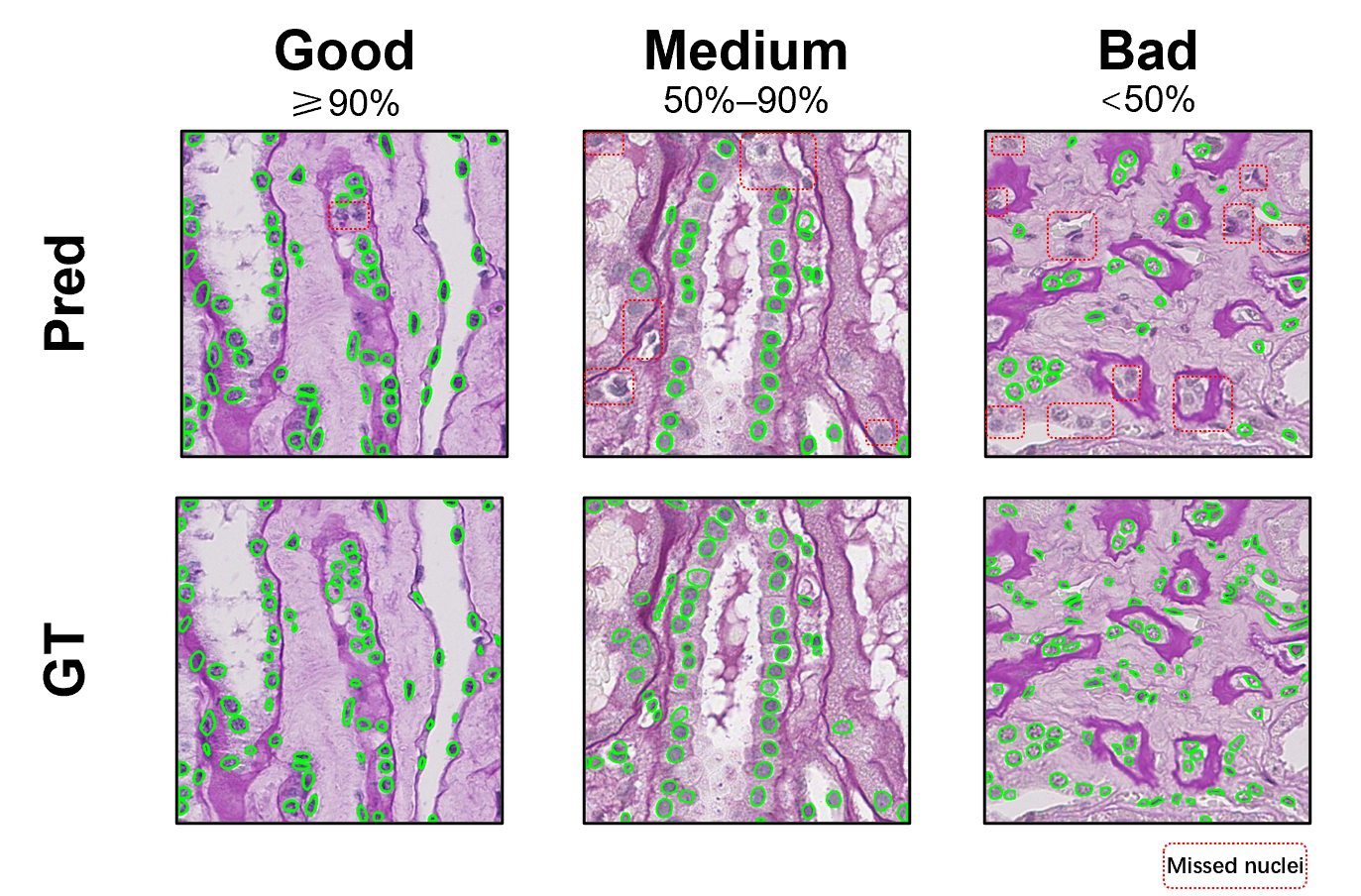} 
    \caption{\textbf{Representative examples of prediction-rating categories.} Foundation-model predictions were categorized as Good, Medium, or Bad according to the proportion of identifiable nuclei captured in each image patch. Green contours denote nucleus instances; the top row shows model predictions, whereas the bottom row shows the corresponding ground-truth annotations.} 
    \label{fig:fig2} 
    \end{figure}

The previous evaluations categorized foundation-model predictions as Good, Medium, or Bad according to the proportion of identifiable nuclei captured in each patch. Predictions capturing at least 90\% of nuclei were rated Good, those capturing approximately 50\%--90\% were rated Medium, and those capturing fewer than 50\% were rated Bad~\cite{guo2025assessment,guo2025evaluating,wang2026evaluating}. Representative examples of the three prediction-rating categories are shown in Fig.~\ref{fig:fig2}.

\begin{figure}[ht]
    \centering
    \includegraphics[width=1\textwidth]{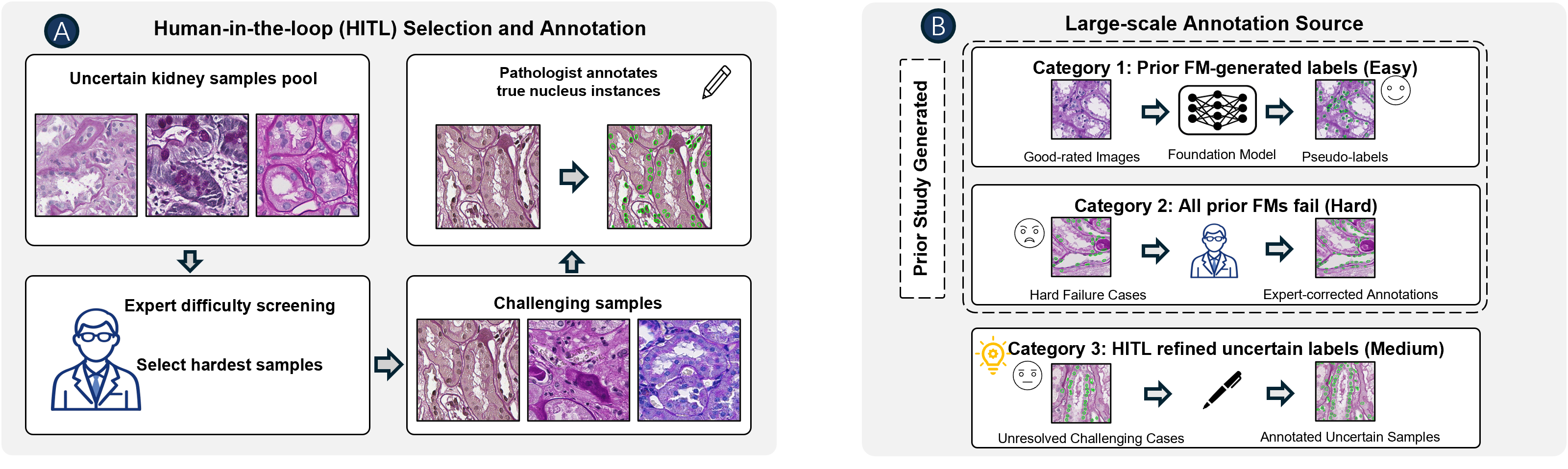}
    \caption{\textbf{Human-in-the-loop annotation and construction of multi-source fine-tuning data.} (A) Pathology experts screened unresolved Medium-quality kidney patches, selected challenging samples, and manually annotated nucleus instances from the original images. (B) The final fine-tuning data consisted of three annotation sources: prior foundation-model-generated pseudo-labels from Easy cases, prior expert-annotated consensus failure cases (Hard), and newly human-in-the-loop annotated unresolved cases (Medium).}
    \label{fig:fig3}
\end{figure}

As shown in Fig.~\ref{fig:fig3}A, pathology experts reviewed the 1,079 unresolved Medium-rated patches and selected 860 of the most challenging cases for manual instance annotation. These samples represented regions that remained insufficiently segmented after the evaluation of newly released models. The human-in-the-loop annotation process was performed from the original image patches rather than by correcting foundation-model predictions.

As shown in Fig.~\ref{fig:fig3}B, the fine-tuning data were organized into three annotation sources according to segmentation difficulty. Easy samples consisted of 5,901 Good-rated image patches with foundation-model-generated pseudo-labels from our prior study. Hard samples consisted of 198 consensus failure cases that were rated Bad across the previously evaluated foundation models and subsequently annotated by pathology experts. Medium samples consisted of the 860 newly selected unresolved challenging patches described above, which were manually annotated from the original images. Together, these three sources provided scalable pseudo-labels, newly annotated unresolved cases, and expert-annotated consensus failures for subsequent fine-tuning experiments.

\subsection{Weighted Oversampling for Annotation Source Rebalancing}

For the four mixed-source datasets---Easy + Medium, Easy + Hard, Medium + Hard, and Easy + Medium + Hard---source-wise weighted oversampling was applied to reduce imbalance among annotation sources, following the rebalancing strategy used in our prior work~\cite{guo2025evaluating}:

\[
w_c = \frac{N}{\gamma n_c + (1-\gamma)N},
\]

where $N$ denotes the total number of training samples, $n_c$ is the number of samples from annotation source $c$, and $\gamma$ controls the strength of source rebalancing. This weighting scheme increased the sampling probability of underrepresented annotation sources and reduced the dominance of the majority source during fine-tuning, thereby increasing model exposure to challenging Medium and Hard patches.

\section{Data and Experiments}

\textbf{Implementations:} The model fine-tuning and inference pipeline was developed using Python 3.12.13 and PyTorch 2.12.1, with CUDA 12.8 enabling GPU acceleration. All experiments were performed on an NVIDIA RTX 5090 GPU with 32 GB of memory. The AdamW optimizer was used for all fine-tuning experiments.

\subsection{Dataset}

\subsubsection{Composition of Fine-tuning Data}

Our previous performance evaluation identified challenging samples that remained unresolved by the cell foundation models~\cite{wang2026evaluating}. Challenging patches were selected in the human-in-the-loop annotation process for fine-tuning cell segmentation models. All identifiable nuclei were annotated as a single foreground class, without nuclear subtype classification. Nuclei intersecting the image boundary were retained, whereas ambiguous objects whose nuclear identity or boundary could not be reliably determined were resolved during expert review. The final annotations were stored as instance-label masks with a unique integer identifier for each nucleus.

As summarized in Table~\ref{tab:annotation_datasets}, three annotation sources were used to construct the fine-tuning datasets: 5,901 Easy pseudo-labels corresponding to the 50\% Easy subset used in our prior study, 860 newly annotated unresolved challenging Medium-quality patches, and 198 prior expert-annotated Hard cases identified as foundation-model failures.~\cite{guo2025evaluating,wang2026evaluating}.

\begin{table}[htbp]
\centering
\caption{Summary of the annotation sources used for model fine-tuning.}
\label{tab:annotation_datasets}

\renewcommand{\arraystretch}{1.45}

\begin{tabularx}{0.90\textwidth}{
    >{\raggedright\arraybackslash}p{2.0cm}
    >{\centering\arraybackslash}p{1.5cm}
    @{\hspace{0.35cm}}
    >{\raggedright\arraybackslash}X
}
\toprule
\textbf{Dataset}
& \textbf{Count}
& \textbf{Annotation Source} \\
\midrule

Easy
& 5,901
& Foundation-model-generated pseudo-labels \\

Medium
& 860
& Unresolved Medium-quality patches annotated by experts \\

Hard
& 198
& Prior foundation-model failure patches annotated by experts \\

\bottomrule
\end{tabularx}
\end{table}

\subsubsection{Test Dataset}
In this study, all experiments were evaluated using the same independent set of 185 expert-annotated kidney pathology patches established in our prior study~\cite{guo2025evaluating}. The test set was excluded from pseudo-label generation, Medium and Hard data enrichment, validation, and fine-tuning.

\subsection{Fine-Tuning Strategies}
As illustrated in Fig.~\ref{fig:fig1}C and summarized in Table~\ref{tab:fine_tuning_strategies}, seven fine-tuning strategies were used: Easy, Medium, Hard, Easy + Medium, Easy + Hard, Medium + Hard, and Easy + Medium + Hard.

The nine cell segmentation model configurations were fine-tuned under all seven strategies, resulting in 63 fine-tuning experiments. As shown in Table~\ref{tab:fine_tuning_strategies}, $\gamma$ was set to 0.85 for Easy + Medium, Easy + Hard, and Easy + Medium + Hard, and to 0.55 for Medium + Hard. These settings reduced dataset imbalance and increased the sampling frequency of the underrepresented Medium and Hard samples. The strategy-specific $\gamma$ values were determined before model fine-tuning based on the degree of dataset imbalance and were applied consistently across all model configurations.

\begin{table}[htbp]
    \centering
    \caption{Fine-tuning data composition and expected sampling ratios for the seven fine-tuning strategies}
    \label{tab:fine_tuning_strategies}

    \small
    \renewcommand{\arraystretch}{1.45}
    \setlength{\tabcolsep}{5pt}

    \begin{tabularx}{\textwidth}{
        @{}
        >{\raggedright\arraybackslash}p{0.24\textwidth}
        >{\centering\arraybackslash}p{0.10\textwidth}
        >{\centering\arraybackslash}p{0.22\textwidth}
        >{\centering\arraybackslash}X
        @{}
    }
        \toprule
        \textbf{Fine-tuning Strategies}
        & \textbf{Samples}
        & \textbf{Original Ratio (\%)}
        & \textbf{Expected Ratio (\%)} \\
        \midrule

        Easy
        & 5,901
        & 100.0
        & 100.0 \\

        Medium
        & 860
        & 100.0
        & 100.0 \\

        Hard
        & 198
        & 100.0
        & 100.0 \\

        Easy + Medium
        & 6,761
        & 87.3 / 12.7
        & 66.5 / 33.5 ($\gamma = 0.85$) \\

        Easy + Hard
        & 6,099
        & 96.8 / 3.2
        & 84.5 / 15.5 ($\gamma = 0.85$) \\

        Medium + Hard
        & 1,058
        & 81.3 / 18.7
        & 72.8 / 27.2 ($\gamma = 0.55$) \\

        Easy + Medium + Hard
        & 6,959
        & 84.8 / 12.4 / 2.8
        & 60.0 / 29.9 / 10.1 ($\gamma = 0.85$) \\

        \bottomrule
    \end{tabularx}

\end{table}

% This setup allows us to directly compare the performance of each standalone model and assess whether ensemble-based approaches enhance segmentation robustness across diverse tissue regions and staining conditions. 

\subsection{Evaluation Metrics}

The performance of all model configurations and their fine-tuned variants was evaluated using the F1 score~\cite{sokolova2009metrics}. Predicted and reference instances were matched using an IoU threshold of 0.5. The F1 score was computed for each test patch and then averaged across all test patches. It was calculated as:

\begin{equation*}
\mathrm{F1} =
\frac{2\mathrm{TP}}
{2\mathrm{TP}+\mathrm{FP}+\mathrm{FN}},
\end{equation*}

where TP, FP, and FN denote true-positive, false-positive, and false-negative nuclei instances, respectively. The F1 score was used as the primary evaluation metric because it jointly accounts for false-positive and false-negative predictions.

\section{Results}

\subsection{Effect of Annotation Strategies Across Model Configurations}
\begin{figure}[!h]
    \centering
    \includegraphics[width=1.0\textwidth]{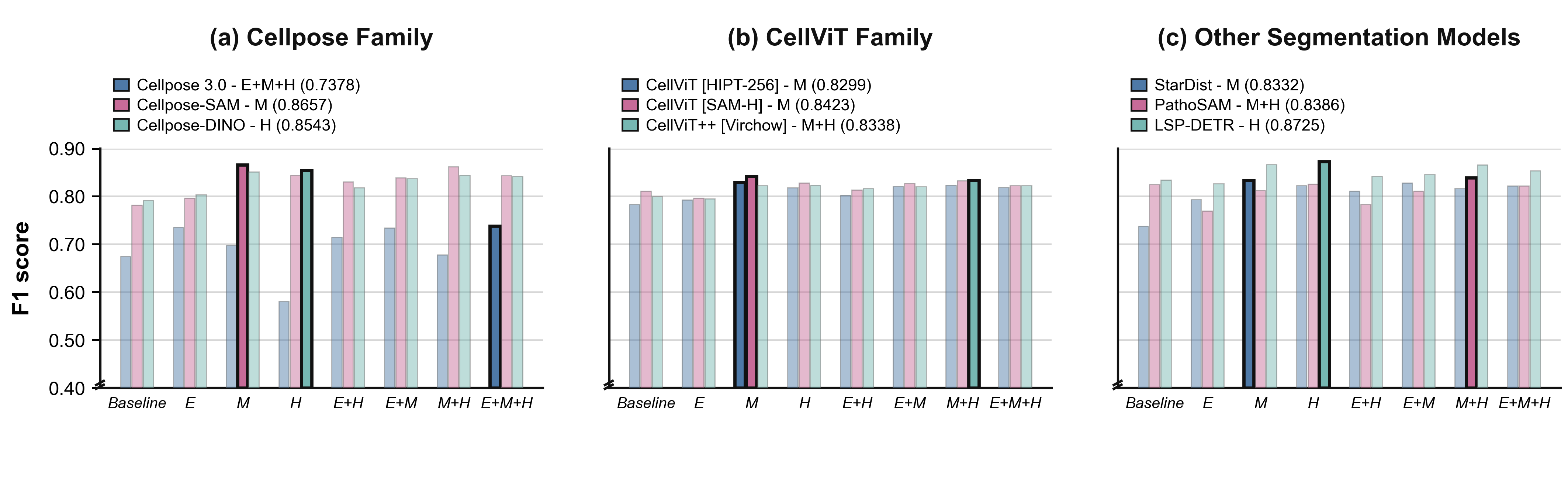} % adjust width & path
    \caption{\textbf{Comparison of F1 scores across seven fine-tuning strategies.} The nine nuclei instance-segmentation model configurations are grouped into (a) the Cellpose family, (b) the CellViT family, and (c) other segmentation models. For each model, performance is shown for the baseline and seven fine-tuning strategies: \textbf{Easy (E)}, \textbf{Medium (M)}, \textbf{Hard (H)}, \textbf{Easy + Hard (E+H)}, \textbf{Easy + Medium (E+M)}, \textbf{Medium + Hard (M+H)} and \textbf{Easy + Medium + Hard (E+M+H)}.}
    \label{fig:f1-strategies}
\end{figure}

In Figure~\ref{fig:f1-strategies}, we compare the F1 scores obtained using the seven annotation strategies across the Cellpose family, CellViT family, and other segmentation models. We found that no single fine-tuning strategy was optimal for all nine model configurations.

Medium-only fine-tuning produced the best F1 score for four models: StarDist, Cellpose-SAM, CellViT [HIPT-256], and CellViT [SAM-H]. We obtained the best results for CellViT++ [Virchow] and PathoSAM by combining Medium and Hard annotations. Cellpose 3.0 achieved its highest F1 score using the complete Easy plus Medium plus Hard dataset. In contrast, Hard-only fine-tuning was optimal for Cellpose-DINO and LSP-DETR.

Overall, seven of the nine best-performing strategies included the newly annotated Medium data. Although Easy + Hard fine-tuning improved performance over the corresponding baseline for eight of the nine model configurations, it was not the optimal strategy for any model once Medium-containing strategies were included in the comparison. These results indicate that the newly annotated Medium cases provided complementary kidney-specific supervision beyond the previously available Easy and Hard data, although the optimal data composition remained model dependent.

\subsection{Overall Fine-Tuning Performance}

\noindent We evaluated all baselines and fine-tuned models on the common 185-patch hold-out test set. Table~\ref{tab:best_strategy_f1} summarizes the best fine-tuning result for each model configuration. We found that fine-tuning improved the best achievable F1 score of all nine model configurations relative to their corresponding baselines. Baseline F1 scores ranged from 0.6748 for Cellpose 3.0 to 0.8339 for LSP-DETR, whereas the best fine-tuned F1 scores ranged from 0.7378 to 0.8725.

Among all configurations, we obtained the highest overall F1 score of 0.8725 with LSP-DETR fine-tuned on the Hard annotations, representing an improvement of 0.0386 over its baseline. Cellpose-SAM achieved the second-highest F1 score of 0.8657 when fine-tuned on the Medium annotations, followed by Cellpose-DINO, which reached 0.8543 using the Hard annotations. Among the CellViT-based configurations, CellViT [SAM-H] achieved the highest F1 score of 0.8423, followed by CellViT++ [Virchow] at 0.8338 and CellViT [HIPT-256] at 0.8299.

We observed substantial differences in the magnitude of improvement across models. StarDist showed the largest absolute gain, increasing from 0.7380 to 0.8332 with Medium-only fine-tuning, corresponding to an improvement of 0.0952. Cellpose-SAM showed the second-largest gain of 0.0841. Cellpose 3.0 and Cellpose-DINO improved by 0.0630 and 0.0625, respectively. We also observed smaller but consistent gains for CellViT [HIPT-256], LSP-DETR, CellViT++ [Virchow], CellViT [SAM-H], and PathoSAM. These results show that the effect of human-in-the-loop data enrichment varied according to the model configuration and its baseline.
\begin{table}[htbp]
    \centering
    \caption{Best fine-tuning strategy and corresponding F1 performance for each model.}
    \label{tab:best_strategy_f1}
    
    \small
    \renewcommand{\arraystretch}{1.35}
    \setlength{\tabcolsep}{6pt}
    
    \begin{tabularx}{\linewidth}{
        @{}
        >{\raggedright\arraybackslash}p{0.25\linewidth}
        >{\centering\arraybackslash}p{0.16\linewidth}
        >{\centering\arraybackslash}p{0.16\linewidth}
        >{\raggedright\arraybackslash}X
        @{}
    }
        \toprule
        \textbf{Cell Segmentation Models} &
        \textbf{Baseline Performance} &
        \textbf{Fine-tuned Performance} &
        \textbf{Fine-tuning Strategy} \\
        \midrule
        
        StarDist
        & 0.7380
        & 0.8332
        & Medium \\
        
        Cellpose 3.0
        & 0.6748
        & 0.7378
        & Easy + Medium + Hard \\
        
        Cellpose-SAM
        & 0.7816
        & \underline{0.8657}
        & Medium \\
        
        CellViT (HIPT-256)
        & 0.7838
        & 0.8299
        & Medium \\
        
        CellViT++ (Virchow)
        & 0.7995
        & 0.8338
        & Medium + Hard \\
        
        CellViT (SAM-H)
        & 0.8108
        & 0.8423
        & Medium \\
        
        Cellpose-DINO
        & 0.7918
        & 0.8543
        & Hard \\
        
        PathoSAM
        & \underline{0.8247}
        & 0.8386
        & Medium + Hard \\
        
        LSP-DETR
        & \textbf{0.8339}
        & \textbf{0.8725}
        & Hard \\
        
        \bottomrule
    \end{tabularx}

    \vspace{2pt}
    \begin{minipage}{\linewidth}
        \footnotesize
        \textit{Note:} Bold values indicate the highest F1 score, and underlined values indicate the second-highest F1 score within each performance column.
    \end{minipage}
\end{table}

\subsection{Qualitative Results }
\begin{figure}[!h]
    \centering
    \includegraphics[width=1.0\textwidth]{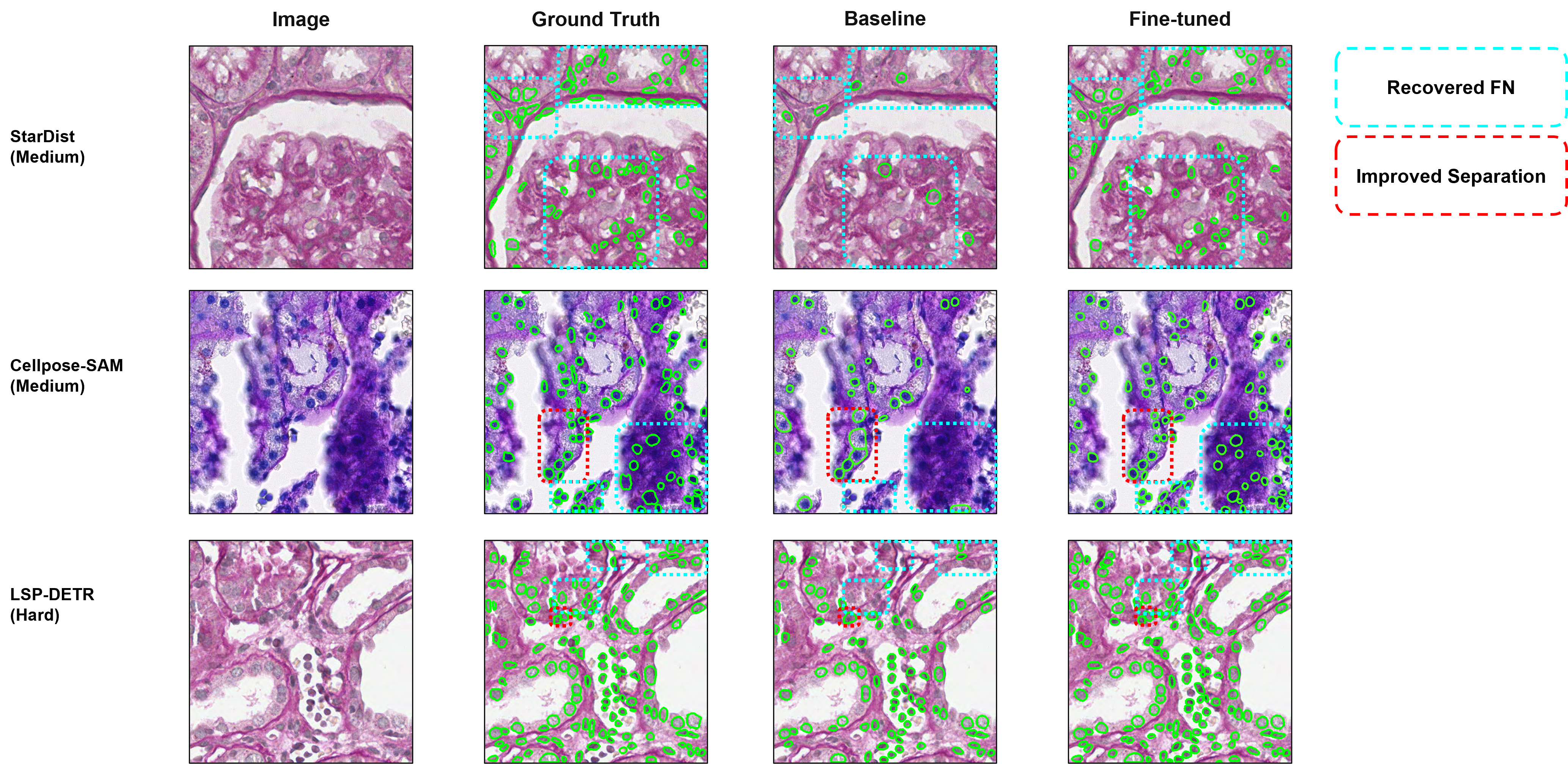} % adjust width & path
    \caption{\textbf{Qualitative comparison of baseline and fine-tuned models.} Green contours denote nuclei instances. Dashed boxes highlight recovered false negatives (FN) and improved instance separation.}
    \label{fig:qualitative-results}
\end{figure}

\noindent In Figure~\ref{fig:qualitative-results}, we present representative test patches comparing the baselines with models fine-tuned using their optimal annotation strategies. We selected examples from StarDist, Cellpose-SAM, and LSP-DETR to illustrate improvements in missed-nucleus recovery and instance separation.

For StarDist, we found that Medium-only fine-tuning recovered multiple nuclei that were missed by baseline, particularly in the highlighted low-contrast and densely populated regions. In the Cellpose-SAM example, we observed improved separation of closely located nuclei together with the recovery of previously missed instances. For LSP-DETR, Hard-only fine-tuning improved the performance of cell nuclei separation while recovering additional nuclei in morphologically complex regions.

These qualitative observations were consistent with the quantitative F1 improvements. We found that the fine-tuned models more accurately captured nuclei in regions with weak contrast, dense nuclear distributions, and complex renal structures. However, differences between the predictions and ground truth remained in some regions, particularly for closely overlapping nuclei and nuclei with poorly defined boundaries.

\section{DISCUSSION}

In this study, we observed substantial variation in the optimal fine-tuning strategies across model configurations, suggesting that the effectiveness of different annotation sources is model dependent. Medium annotations were included in the best-performing strategies for seven of the nine model configurations, indicating that expert annotation of previously unresolved cases provides useful supervision beyond large-scale Easy pseudo-labels and a small set of Hard consensus failures. Although Easy + Hard fine-tuning improved over the baseline for eight of the nine models, it was not optimal for any model once Medium-containing strategies were considered, suggesting that the newly annotated Medium cases provided complementary supervision beyond the data used in our prior human-in-the-loop study. In particular, Medium-only fine-tuning achieved the best performance for StarDist, Cellpose-SAM, and both CellViT configurations, whereas Medium + Hard was optimal for CellViT++ [Virchow] and PathoSAM. However, Medium data were not universally required: Cellpose-DINO and LSP-DETR achieved their best performance using Hard-only fine-tuning, while Cellpose 3.0 benefited most from the complete Easy + Medium + Hard dataset. These differences suggest that the value of each annotation source may depend on model-specific characteristics, including pretrained representations, segmentation formulations, and baseline alignment with renal pathology. Within the Cellpose family, this behavior may be related to the combination of Cellpose 3.0's conventional CNN encoder and the grayscale, DAPI-like preprocessing used in our pipeline, which may reduce discriminative information in faintly stained and low-contrast Medium and Hard cases. In contrast, Cellpose-SAM and Cellpose-DINO leverage large-scale pretrained ViT encoders and may therefore be more robust to these challenging appearances.

The frequent benefit of Medium annotations also suggests that samples remaining unresolved after successive foundation-model evaluations contain complementary information that is not fully represented by either well-segmented Easy cases or extreme Hard failures. However, because no single annotation composition consistently performed best across all models, a fixed fine-tuning dataset may not be optimal for adapting diverse segmentation architectures. Several limitations should also be considered. First, this study focuses exclusively on renal pathology, and the effectiveness of the proposed annotation strategies may differ across organs, staining protocols, or imaging domains. Second, evaluation was performed on a fixed hold-out set of 185 expert-annotated kidney pathology patches. In addition, the source-rebalancing parameters were predefined for each mixed-data strategy and applied consistently across models rather than optimized separately for individual architectures. Finally, each model–strategy configuration was evaluated using a single fine-tuning run, and variability arising from random initialization or stochastic optimization was not explicitly quantified. Future work could investigate model-specific or adaptive data-selection strategies that automatically identify the most informative annotation sources and determine how annotation composition should be adjusted for different segmentation models.

\section{Conclusion}

In this study, we evaluated seven single-source and mixed-source fine-tuning strategies using Easy pseudo-labels, newly annotated Medium cases, and expert-annotated Hard consensus failures across nine cell nuclei segmentation model configurations. For every configuration, at least one fine-tuning strategy improved F1 over its corresponding baseline on the independent kidney nuclei test set. LSP-DETR achieved the highest F1 score of 0.8725 with Hard-only fine-tuning, while StarDist showed the largest improvement, increasing from 0.7380 to 0.8332 with Medium-only fine-tuning. Medium data were included in seven of the nine optimal strategies, demonstrating the value of targeted annotation of previously unresolved cases. However, no single data composition was optimal across all models. These findings support human-in-the-loop annotation enrichment across different segmentation-difficulty levels as an effective approach for adapting general-purpose nuclei segmentation models to renal pathology.

% \acknowledgements
% % equivalent to \section*{ACKNOWLEDGMENTS}       
% \begin{flushleft}
% This research was supported by NIH R01DK135597 (Huo), DoD HT9425-23-1-0003 (HCY), NSF 2434229 (Huo) and KPMP Glue Grant. This work was also supported by Vanderbilt Seed Success Grant and Vanderbilt-Liverpool Seed Grant. This research was also supported by NIH grants R01EB033385, R01DK132338. We extend gratitude to NVIDIA for their support by means of the NVIDIA hardware grant. 

% \end{flushleft}
\bibliography{report}
\bibliographystyle{spiebib}

\end{document}